# Automatically Detecting Self-Reported Birth Defect Outcomes on Twitter for Large-scale Epidemiological Research

Ari Z. Klein[1], Abeed Sarker[1], Davy Weissenbacher[1], Graciela Gonzalez-Hernandez[1]

[1] Department of Biostatistics, Epidemiology, and Informatics, Perelman School of Medicine, University of Pennsylvania, Philadelphia, PA, United States

{ariklein, abeed, dweissen, gragon}@pennmedicine.upenn.edu

**ABSTRACT**

*Background*: In recent work, we identified and studied a small cohort of Twitter users whose pregnancies with birth defect outcomes could be observed via their publicly available tweets. Exploiting social media's large-scale potential to complement the limited methods for studying birth defects—the leading cause of infant mortality—depends on the further development of automatic methods.
*Objective*: The primary objective of this study was to take the first step towards scaling the use of social media for observing pregnancies with birth defect outcomes—namely, developing methods for automatically detecting tweets by users reporting their birth defect outcomes.
*Methods*: We annotated and pre-processed approximately 23,000 tweets that mention birth defects in order to train and evaluate supervised machine learning algorithms, including feature-engineered and deep learning-based classifiers. We also experimented with various under-sampling and over-sampling approaches to address the class imbalance.
*Results*: A Support Vector Machine (SVM) classifier trained on the original, imbalanced data set, with n-grams, word clusters, and structural features, achieved the best baseline performance for the positive classes: an $F_1$-score of 0.65 for the "defect" class and 0.51 for the "possible defect" class.
*Conclusions*: Our contributions include (i) natural language processing (NLP) and supervised machine learning methods for automatically detecting tweets by users reporting their birth defect outcomes, a (ii) a comparison of feature-engineered and deep learning-based classifiers trained on imbalanced, under-sampled, and over-sampled data, and (iii) an error analysis that could inform classification improvements using our publicly available corpus. Future work will focus on automating user-level analyses for cohort inclusion.

## 1. Background and Significance

Considering that birth defects are leading cause of infant mortality in the United States [1], methods for observing pregnancies with birth defect outcomes remain limited (e.g., clinical trials [2,3], animal studies [4,5], pregnancy exposure registries [5-8]), and 36% of Americans between ages 18-29 use Twitter [9], in recent work [10], we used natural language processing (NLP) and manual analysis methods to identify a cohort of women whose pregnancies with birth defect outcomes could be observed via their publicly available tweets. Then, in a pharmacoepidemiological case-control study [11], we compared risk factors [12] reported by the identified cohort and an internal comparator group selected from the same database, which stores the *timelines*—the publicly available tweets posted over time—of users automatically identified via their public announcements of pregnancy on Twitter [13]. We found complementary evidence on social media that taking medication [14] during pregnancy is associated with a higher risk of birth defect outcomes.

While promising, our preliminary studies also demonstrate the limitations of *manual* methods for identifying a cohort on social media. Upon grappling with collecting sparse tweets for manual annotation, we distinguished true and false positives among approximately 17,000 tweets that mention birth defects, and then manually analyzed approximately 650 timelines of users who posted true positives. In the end, we identified only 195 women for inclusion in our cohort [10], which prevented our case-control study [11] from focusing on individual birth defects. Despite the relatively small cohort initially identified and studied, additional users are being constantly added to our database [13] over time. Exploiting social media on a scale potentially large enough for studying individual birth defects—most of which have unknown etiologies [15]—depends on methods capable of *automatically* identifying cohorts. However, considering that the prevalence of birth defect outcomes in the United States is only 3% [16] (even postpartum depression [17] affects up to 15% of mothers [18]), automatically detecting such rare events on social media presents significant challenges [19].

Tweets that mention birth defects are extremely sparse, and approximately 90% of the ones we retrieved are false positives, so we spent more than a year collecting and annotating real-time Twitter data in order to develop a training set for automatically distinguishing tweets (possibly) indicating that the user's child has a birth defect (true positives). The high degree of class imbalance represented in the training data makes automating this distinction a difficult classification task for supervised machine learning [20-22]. The primary objective of this study is to assess the utility of our annotated data as a training set for *automatically* detecting tweets by mothers reporting that their child has a birth defect. This classification, in turn, would be the first step towards automatically identifying pregnancies with birth defect outcomes that can be observed on Twitter for large-scale epidemiological research. In this paper, we present (i) baseline results of NLP and supervised classification methods, (ii) a comparison of feature-engineered and deep learning-based classifiers trained on imbalanced, under-sampled, and over-sampled data, and (iii) an error analysis that could inform strategies for improving classification performance using the annotated corpus we have made publicly available [23].

## 2. Methods

### 2.1. Data Collection

To collect tweets for training and evaluating supervised machine learning algorithms, we first compiled a lexicon of birth defects [24-28]; the Penn Social Media Lexicon of Birth Defects consists of approximately 650 single-word and multi-word terms. To enhance retrieval in social media, we semi-automatically generated lexical variants of these terms (e.g., misspellings, inflections) using an automatic, data-centric spelling variant generator [29]. The Penn Social Media Lexicon of Birth Defects and its lexical variants are available in Supplementary Material of our related work [10]. To retrieve tweets containing (variants of) the terms, we implemented hand-crafted regular expressions for mining more than 400 million tweets, posted by more than 100,000 users, stored in our database [13]. We post-processed the matching tweets by removing retweets and tweets containing user names and URLs that were matched by the regular expressions. Using the current version of the query, we have retrieved a total of 23,680 tweets, including the tweets retrieved for our feasibility study [10], which describes our data collection methods in more detail.

## 2.2. Annotation

Two professionally trained annotators annotated 23,680 tweets, posted by 7,797 Twitter users, with overlapping annotations for 22,408 (94.63%) tweets. Annotation guidelines, available in Supplementary Material of our related work [10], were developed to help the annotators distinguish three classes of tweets: "defect," "possible defect," and "non-defect." Table 4 (in Section 4.2) includes examples of "defect" and "possible defect" tweets, and our feasibility study [10] provides additional examples and a discussion of the three classes, summarized as follows:

- *Defect* (+): The tweet refers to a person who has a birth defect and identifies that person as the Twitter user's child.
- *Possible Defect* (?): The tweet is ambiguous about whether a person referred to has a birth defect and/or is the Twitter user's child.
- *Non-defect* (-): The tweet does not indicate that a person referred to has or may have a birth defect and is or may be the user's child.

Inter-annotator agreement was $\kappa = 0.86$ (Cohen's kappa), considered "almost perfect agreement" [30]. The tweets on which the annotators disagreed were excluded from the final data set, which consists of 22,999 annotated tweets: 1,192 (5.12%) "defect" tweets, 1,196 (5.20%) "possible defect" tweets, and 20,611 (89.67%) "non-defect" tweets. The "possible defect" class is a "placeholder" class indicating that analysis of the user's timeline is needed for contextually determining if they are the mother of a child with a birth defect. In our feasibility study [10], we found that approximately 40% of the users who posted a "possible defect" tweet (without also posting a "defect" tweet) indeed had a child with a birth defect. Automating the analysis of user timelines, however, is beyond the scope of this paper and will be explored in future work. For now, we focus on the task of automatically classifying tweets, as described next.

## 2.3. Classification

We used the data set of 22,999 annotated tweets in experiments to train and evaluate supervised machine learning algorithms. The goal of these experiments was to generate baseline performance results and assess the utility of the annotated corpus as a training set for automatic classification. For the classifiers, we used (i) the default implementation of Naïve Bayes (NB) in Weka 3.8.2, (ii) the WLSVM Weka integration [31] of the LibSVM [32] implementation of Support Vector Machine (SVM), and (iii) a Bidirectional Long Short-Term Memory Unit Deep Recurrent Neural Network (Bi-LSTM) [33,34], implemented in Keras running on top of TensorFlow. We split the data into 80% (training) and 20% (test) stratified random sets. To optimize training, we further split the training set into 80% (training) and 20% (validation) stratified random sets. Thus, our training set consists of 14,716 tweets, our validation set consists of 3,681 tweets, and our test set consists of 4,602 tweets. In the remainder of this sub-section, we will describe the data pre-processing, feature sets, classifier parameters, and data imbalance approaches used in supervised classification.

### 2.3.1. Pre-processing and Features

For the NB and SVM classifiers, we performed standard text pre-processing by lowercasing and stemming [35] the tweets. We also removed non-alphabetical characters (e.g., hashtags, ellipses, UTF-8 special characters), and normalized user names (i.e., strings beginning with "@") and URLs. We did not remove stop words because, based on a preliminary feature evaluation using *information gain* [36], we found that stop words (e.g., *my*, *she*, *was*, *with*, *have*) occur in n-grams—contiguous sequences of *n* words—that are highly informative for distinguishing the three classes (e.g., *my baby*, *daughter with*, *she*, *was born*, *may have*). We also noticed in the feature ranking that some of the n-grams were semantically similar to n-grams that were not ranked as high (e.g., *our son*, *child with*, *he*). Given the context of this data set, we assumed that the hierarchy of these n-grams does not represent a meaningful semantic distinction but is merely a byproduct of the various linguistic ways in which users are referring to their children; thus, we reduced the semantic feature space by normalizing particular first-person pronouns (e.g., *my*, *our*), references to a child (e.g., *son*, *daughter*, *child*), and third-person pronouns (e.g., *she*, *he*), respectively.

The information gain feature evaluation also suggested that, in the raw data, specific birth defect terms were informative linguistic features for distinguishing the classes. As Table 1 shows, such terms were mentioned markedly more frequently in "non-defect" tweets, suggesting that they tend to play a role in linguistically characterizing the "non-defect" class—an artifact of the imbalanced data set, but not an accurate representation of human decision making for a machine learning model. In addition, the birth defect terms in Table 1 represent some of the more frequently mentioned birth defects on Twitter, while most of the others are relatively sparse. In order to avoid overfitting the machine learning model to birth defects that are mentioned relatively frequently and tend to accompany "non-defect" tweets, we also normalized the span of the tweet that was matched by the regular expressions in the database query [10].

**Table 1**. Examples and per-class frequencies of birth defect terms that distinguish the "defect" (+), "possible defect" (?), and "non-defect" (-) tweet classes in the raw annotated data.

| Birth Defect Term | + | ? | - |
|---|---|---|---|
| CHD | 115 | 79 | 914 |
| Club Foot | 65 | 35 | 207 |
| Down Syndrome | 407 | 467 | 11,157 |
| Dwarfism | 18 | 18 | 381 |
| Gastroschisis | 27 | 23 | 100 |
| Hydrocephalus | 22 | 39 | 256 |
| Microcephaly | 54 | 11 | 597 |
| Trisomy 18 | 82 | 38 | 232 |

Finally, we also normalized people's names in pre-processing the tweets. During manual annotation, we found that, oftentimes, "possible defect" tweets are characterized by an ambiguous referent of a person's name; that is, we do not know if the name refers to the user's child. However, because of the heterogeneity of specific names across tweets, names would not be modeled as (part of) salient n-grams. To detect names in the tweets, we compiled a lexicon of the 200 most popular boys names and the top 200 girls names in the United States between 2010 and the present, identified by the Social Security Administration [37]. Following pre-processing, for the NB and SVM classifiers, we converted the tweets to word vectors and used Weka's default N-Gram Tokenizer to extract unigrams (n = 1), bigrams (n = 2), and trigrams (n = 3) as features.

In addition to n-grams, we explored word clusters and structural features for the SVM classifier. Word clusters provide generalized representations of semantically similar terms, in that terms appearing in similar collocate contexts, such as misspellings, are represented by the same cluster. The clusters are generated via a two-step process: first, vector representations of the words (i.e., word embeddings) are learned from large, unlabeled social media data, and then, the vectors are grouped via a standard clustering algorithm. We have found generalized representations via cluster numbers to be useful in our past work [38]. For the present work, we used the Twitter word clusters made publicly available by Owoputi et al. [39]. For structural features, we used the tweet's length in characters and words.

For the Bi-LSTM classifier, features were learned using the publicly available GloVe word vectors trained on two billion tweets [40]. We limited the size of the vectors to 100 dimensions. Any word in our corpus not found in the vocabulary of the pre-trained vectors was represented with a zero vector and updated during training. We pre-processed the raw tweets in our corpus with the same operations and sequence used when creating the vectors. Specifically, we first normalized all user names (<USER>) and URLs (<URL>). Next, we inserted spaces before and after each slash character, and normalized all numbers (<NUMBER>). Then, we normalized all repetitions of the same punctuation (<REPEAT>) following the occurrence of the first one. Similarly, we deleted all letters repeated more than three times, and marked the occurrence of elongated words (<ELONG>). Finally, we replaced hashtag (#) characters (<HASHTAG>). We tokenized the tweets with the PTBTokenizer, and lowercased the tokens.

### 2.3.2. Classifier Parameters

For the SVM classifier [41], we used the radial basis function (RBF) kernel and, upon tuning, set the cost value at c = 100. To address the class imbalance problem, which we will discuss in more detail in Section 2.3.3, we assigned higher weights to the minority classes. We scaled the feature vectors before applying the SVM for classification. For the Bi-LSTM classifier, the neural network consists of three layers sequentially connected: (i) the input layer that passes the word embedding vectors to the second layer, (ii) an LSTM layer with 128 dimensions and a tanh activation function, and (iii) a fully connected layer with three dimensions and a softmax function. We applied a dropout function with a probability of 0.5 during training. We trained the Bi-LSTM classifier for 15 iterations.

### 2.3.3. Data-Level Approaches to Class Imbalance

In addition to assigning higher weights to the minority classes with the SVM classifier, we performed under-sampling of the majority ("non-defect") class using several methods to address the class imbalance problem at the data level: (i) removal of majority class instances that were lexically similar to other instances in the training data, (ii) removal of majority class instances that were lexically similar to minority class instances in the validation set that were being misclassified as "non-defect" in the preliminary experiments, and (iii) random under-sampling of the majority class. During annotation, we found a number of cases in which the same or a similar "non-defect" tweet was posted numerous times (e.g., fundraisers, news headlines, advertisements). Thus, our assumption underlying method (i) was that we could reduce the imbalance without losing linguistic information about the majority class. For method (ii), we attempted to strategically remove "non-defect" tweets that were close to the linguistic boundary of the minority classes.

To compute lexical similarities between tweets, we employed the *Levenshtein ratio* (LR) method, which is calculated as $LR = \frac{(lensum - lendist)}{lensum}$, where *lensum* is the sum of the lengths of the two strings, and *lendist* is the Levenshtein distance between the two strings. The equation results in a value between 0 (completely dissimilar strings) and 1 (identical strings). Levenshtein distance is a measure of the similarity between two strings computed as the number of deletions, insertions, or substitutions required to transform one string into another. When performing the under-sampling, we removed tweets that produced LR scores above a pre-defined threshold ($k$). We were able to choose the size of the under-sampled training sets by varying the value of $k$. We controlled methods (i) and (ii)—the similarity-based under-sampling methods—by employing method (iii)—random under-sampling of the majority class—based on the final size of the training sets for methods (i) and (ii).

We also performed over-sampling of the minority ("defect" and "possible defect") classes. We balanced the training set by (iv) over-sampling the minority classes with replacement (i.e., multiplying the original sets of "defect" and "possible defect" tweets until they were nearly equal to the number of "non-defect" tweets), and (v) utilizing the Synthetic Minority Over-sampling Technique (SMOTE) [42]. With SMOTE, we nearly balanced the training set by creating artificial instances of "defect" and "possible defect" tweets in vector space. We used the default settings for the SMOTE filter in Weka. We performed all of the other over-sampling and under-sampling methods for both the SVM and Bi-LSTM classifiers, but applied SMOTE only for the SVM classifier.

### 3. Results

Figure 1 presents the performance of the NB, SVM, and Bi-LSTM classifiers, as measured by $F_1$-score, for the "defect" (+), "possible defect" (?), and "non-defect" (-) classes. $F_1$-score is computed as *2 x (Precision x Recall)/(Precision + Recall)*, where precision is *True Positives/(True Positives + False Positives)*, and recall is *True Positives/(True Positives + False Negatives)*. In Figure 1, the classifiers were trained on the original, imbalanced data set, with only n-grams as features for the NB and SVM classifiers. The SVM and Bi-LSTM classifiers outperformed the NB baseline for all three classes, and the SVM classifier outperformed the Bi-LSTM classifier for the "defect" and "possible defect" classes, and achieved similar performance for the "non-defect" class.

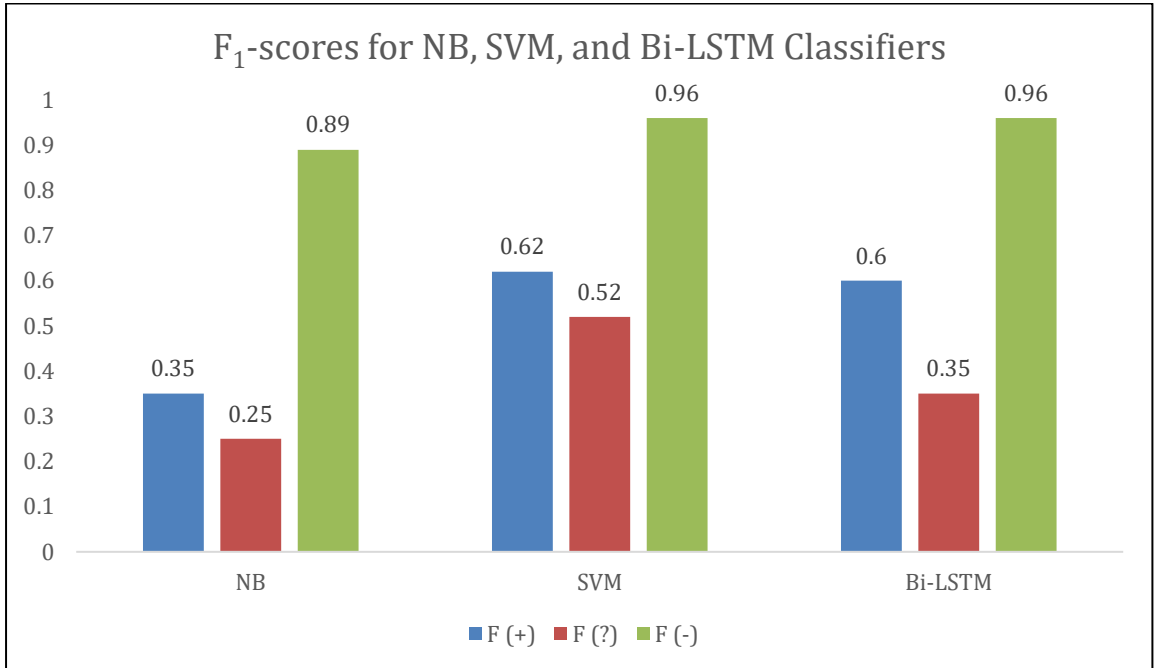

**Figure 1**. $F_1$-scores (F) for "defect" (+), "possible defect" (?), and "non-defect" (-) tweet classes, for Naïve Bayes (NB), Support Vector Machine (SVM), and Deep Recurrent Neural Network (Bi-LSTM) classifiers trained on the original, imbalanced data set, with n-grams as features for the NB and SVM classifiers.

To address the class imbalance problem at the data level, the SVM and Bi-LSTM classifiers were trained on variations of the original, imbalanced data, as described in Section 2.3.3. Table 2 compares the performance of the SVM and Bi-LSTM classifiers when they were trained on the original data with their performance when they were trained using various under-sampling and over-sampling approaches. For the SVM classifier, neither under-sampling nor over-sampling improved performance for any of the classes. For the Bi-LSTM classifier, neither under-sampling nor over-sampling improved performance for the "defect" (+) or "non-defect" (-) classes, but most of the approaches did improve performance for the "possible defect" (?) class; however, the similarity-based under-sampling methods did not outperform their random under-sampling controls. Overall, though, the Bi-LSTM did not outperform the SVM, so we decided to pursue the SVM that was trained on the original data set.

**Table 2**. Overall and per-class $F_1$-scores (F) for "defect" (+), "possible defect" (?), and "non-defect" (-) tweet classes, for Support Vector Machine (SVM) and Deep Recurrent Neural Network (Bi-LSTM) classifiers trained on the original, under-sampled, and over-sampled data sets, with n-grams as features for the SVM classifier.

| Classifier | Training Set | F (+) | F (?) | F (-) | F |
|---|---|---|---|---|---|
| SVM | original, imbalanced training set (14,716) | 0.62 | 0.52 | 0.96 | 0.92 |
| SVM | under-sampling based on similar majority class tweets in original training set (5,551)[a] | 0.62 | 0.43 | 0.96 | 0.91 |
| SVM | random under-sampling control set (5,551)[b] | 0.62 | 0.49 | 0.96 | 0.92 |
| SVM | under-sampling based on similar false negative majority class tweets (8,015)[c] | 0.58 | 0.51 | 0.95 | 0.91 |
| SVM | random under-sampling control set (8,015)[b] | 0.62 | 0.50 | 0.96 | 0.91 |
| SVM | over-sampling instances of minority classes with replacement (40,675)[d] | 0.62 | 0.46 | 0.95 | 0.91 |
| SVM | SMOTE on original training set (39,148)[e] | 0.62 | 0.51 | 0.96 | 0.92 |
| Bi-LSTM | original, imbalanced training set (14,716) | 0.60 | 0.35 | 0.96 | 0.91 |
| Bi-LSTM | under-sampling based on similar majority class tweets in original training set (5,551)[a] | 0.55 | 0.33 | 0.91 | 0.86 |
| Bi-LSTM | random under-sampling control set (5,551)[b] | 0.54 | 0.37 | 0.92 | 0.87 |
| Bi-LSTM | under-sampling based on similar false negative majority class tweets (8,015)[c] | 0.48 | 0.36 | 0.90 | 0.85 |
| Bi-LSTM | random under-sampling control (8,015)[b] | 0.59 | 0.45 | 0.95 | 0.91 |
| Bi-LSTM | over-sampling instances of minority classes with replacement (40,675)[d] | 0.55 | 0.45 | 0.95 | 0.91 |

[a] Method (i) described in Section 2.3.3.
[b] Method (iii) described Section 2.3.3.
[c] Method (ii) described in Section 2.3.3.
[d] Method(iv) described in Section 2.3.3.
[e] Method (v) described in Section 2.3.3.

In addition to n-grams, we explored additional features for the SVM classifier, as described in Section 2.3.1. Table 3 presents the precision, recall, and $F_1$-score of the classifier with the addition of word clusters and structural features to the standalone n-grams set. Based on the default implementation of a paired t-test in Weka's Experiment Environment, the additional features significantly improved the $F_1$-score for the "defect" class ($P<0.05$). The results of leave-one-out classification experiments in Table 3 indicate, however, that the structural features did not contribute to the improved performance. Thus, word clusters, in particular, increased the classifier's overall performance, though increasing recall at the expense of precision. Although the addition of word clusters significantly improved the classifier's performance, the results of (i) using word clusters as a standalone feature set or (ii) leaving out n-grams in feature ablation significantly decreased the $F_1$-score for all three classes ($P<0.05$), indicating, unsurprisingly, that the classifier's performance can be attributed primarily to the information provided by n-grams.

Table 3. Precision (P), recall (R), and $F_1$-scores (F) for Support Vector Machine (SVM) classifier for "defect" (+), "possible defect" (?), and "non-defect" (-) tweet classes, for leave-one-out and standalone feature classification.

| Features | P (+) | R (+) | F (+) | P (?) | R (?) | F (?) | P (-) | R (-) | F (-) |
|---|---|---|---|---|---|---|---|---|---|
| All | 0.62 | 0.68 | 0.65 | 0.58 | 0.45 | 0.51 | 0.96 | 0.96 | 0.96 |
| w/o N-Grams | 0.22 | 0.55 | 0.32 | 0.43 | 0.22 | 0.29 | 0.94 | 0.89 | 0.92 |
| w/o Word Clusters | 0.67 | 0.58 | 0.62 | 0.58 | 0.46 | 0.52 | 0.95 | 0.97 | 0.96 |
| w/o Structural Features | 0.62 | 0.68 | 0.65 | 0.57 | 0.45 | 0.51 | 0.96 | 0.96 | 0.96 |
| N-Grams | 0.67 | 0.58 | 0.62 | 0.58 | 0.46 | 0.52 | 0.95 | 0.97 | 0.96 |
| Word Clusters | 0.20 | 0.58 | 0.30 | 0.43 | 0.22 | 0.29 | 0.95 | 0.87 | 0.91 |

## 4. Discussion

### 4.1. Principal Results

The baseline results of classification suggest that our annotated data has utility for training machine learning algorithms to automatically detect tweets by mothers (possibly) reporting that their child has a birth defect. This automatic classification enables the long-term collection of social media data on a scale potentially large enough for studying individual birth defects, such as congenital heart defects (CHDs), which are the most common type of birth defect in the general population [43] and reported on Twitter [10], are the leading cause of infant mortality due to birth defects [44], and have a largely unknown etiology [45]. The comparative performance of the SVM classifier is consistent with our recent findings that deep neural network-based classifiers are generally outperformed for imbalanced social media data [46]. Our results also suggest that data-level approaches for addressing class imbalance with convolutional neural networks [47] may not generalize to recurrent neural networks (e.g., LSTMs).

The $F_1$-scores for the minority classes of the SVM classifier (0.62 for "defect" and 0.51 for "possible defect" tweets), while promising baselines, are relatively low. The high $F_1$-score for the "non-defect" class (0.96) could be attributed to the class imbalance when trained on the original data, but the $F_1$-score remained relatively high when the imbalance was reduced (under-sampling) or nearly eliminated (over-sampling). Thus, the relatively low performance for the minority classes could be due to the broad linguistic scope represented by the "non-defect" class—that is, the wide variety of linguistic patterns surrounding mere mentions of birth defects. Because these linguistic patterns are also likely to occur in "defect" and "possible defect" tweets, the classifier could confuse them as "non-defect," especially if the "defect" or "possible defect" tweets do not explicitly encode that the user's child (possibly) has a birth defect. We will provide examples of such tweets in the error analysis that follows.

### 4.2. Error Analysis

We conducted a brief error analysis of the SVM classifier trained on the original training set with n-grams, word clusters, and structural features, focusing on false negative "defect" and "possible defect" tweets—more specifically, "defect" and "possible defect" tweets misclassified as "non-defect." Our analysis of the false negative "possible defect" tweets—the classifier's primary confusion—reveals that the majority of them contain the name of a person. For example, Tweet 1 in Table 4 was annotated as "possible defect" because it is ambiguous as to whether the referent of the person is the user's child. As described Section 2.3.1, we attempted to model names as a normalized representation in pre-processing; however, our analysis reveals that more than half of the names were not matched by the lexicon. Many of the other false negative "possible defect" tweets contain a variety of such references, such as *he* in 2 and *this girl* in 3, or they omit an explicit lexical reference entirely, as in 4.

**Table 4**. Samples of false negative "defect" (+) and "possible defect" (?) tweets, misclassified as "non-defect" (-) by the SVM classifier trained on the original data set, with n-grams, word clusters, and structural features.

|    | Tweet | Actual | Predicted |
|----|-------|--------|-----------|
| 1  | [name] was diagnosed with craniosynostosis and had surgery to repair it. | ? | - |
| 2  | He has a cleft palate.....I don't think he's special needs. | ? | - |
| 3  | I'm just in love w. this little boy. Hydrocephalus, split cerebellum, 2 vessel cord, dilated kidney, & survived | ? | - |
| 4  | Born @ 25 weeks, hole in heart, no anal area, narrow airway, no vocal chords, floppy voicebox, missing vertebrae. Healed! | ? | - |
| 5  | They couldn't get a good picture of 👶 foot, and DR said it could be a Club Foot. | ? | - |
| 6  | #DUPCoalition say the decision I was forced to make for my unborn daughter was wrong Trisomy 13, alobar holoprosencephaly,cystic hygroma | + | - |
| 7  | Her face 😊#lovemygirl #downsyndrome #motheranddaughter 👤👶 | + | - |
| 8  | Raising a daughter with Down syndrome makes me dream of a more inclusive society | + | - |
| 9  | My #clubfoot #cutie 💚👣 | + | - |
| 10 | We were given no options other than termination "We have no resources for you" 😞 #Trisomy18 | + | - |

Tweet 1 explicitly states that the person referred to by name has craniosynostosis (*was diagnosed with*), and 2 explicitly states that *he* has a cleft palate (*has*), but most of the false negative "possible defect" tweets do not as explicitly encode this information in the text, as in 3 and 4. A human can reasonably infer that *this little boy* in 3 *has* hydrocephalus, and that the child implicitly referred to in 4 *was born with* a hole in her heart, but, without explicitly encoding this relationship between the person and the birth defect, the textual surface of the tweets may more closely resemble "non-defect" tweets that merely mention birth defects. Similarly, some of the false negative "possible defect" tweets do not explicitly state that the child has a birth defect because, as in 5, they are ambiguous as to whether the child actually has the birth defect (*could be a club foot*). Figures 2 summarizes our analysis of errors for the false negative (FN) "possible defect" (?) and "defect" (+) tweets in our held-out test set. As the proportions of errors indicate, many of the tweets contain multiple possible sources of error.

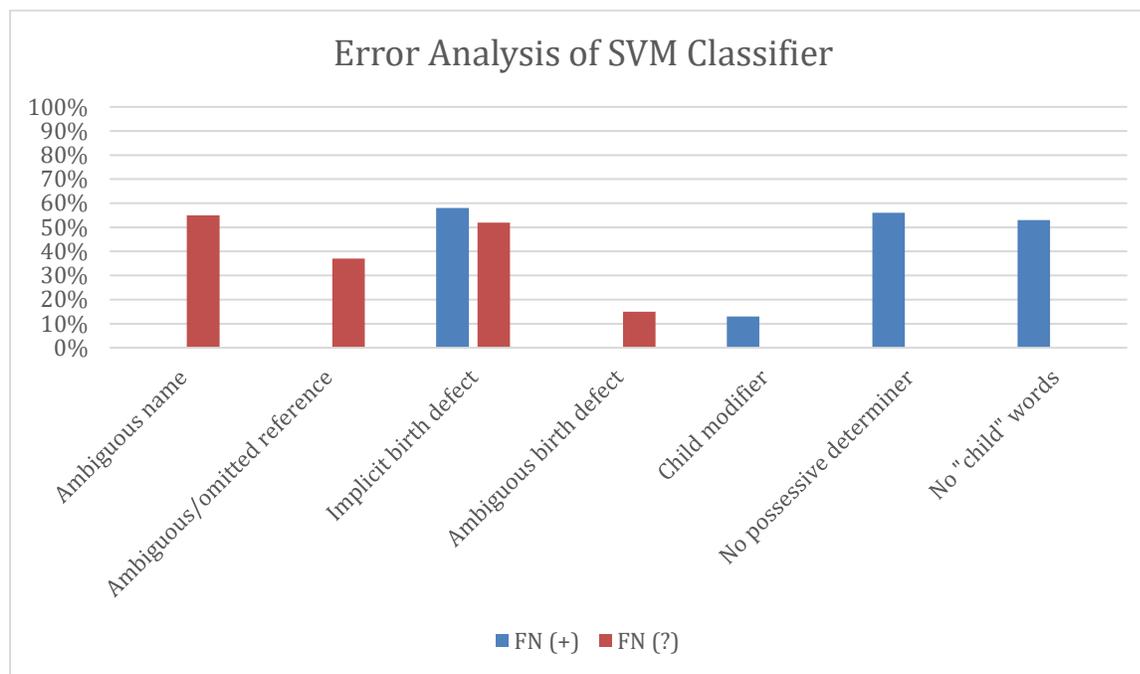

**Figure 2**. Possible sources of error for false negative (FN) "defect" (+) and "possible defect" (?) tweets predicted by the SVM classifier trained on the original data set, with n-grams, word clusters, and structural features.

Most of the false negative "defect" tweets also imply having a birth defect. In addition, many of them, while referring to the user's child, do not contain a common reference, such as *my child*. Some of the tweets, such as 6, explicitly refer to the user's child (*my daughter*), but obscure this "defect" information by interrupting the pattern with a modifier (*my unborn daughter*). Many of the false negatives do not contain a possessive determiner (e.g., *my*, *our*) explicitly indicating that the user is referring to *her* child. Although 7 does contain the uninterrupted pattern, *my girl*, it is textually represented as a hashtag and, thus, modeled as part of a single token: *lovemygirl*. In 8, the first-person pronoun, *me*, suggests that the user is referring to her daughter, but *daughter* itself occurs with a determiner, *a*, that does not express a reference to the user's child. Furthermore, many of the tweets do not use common "child" words. For example, 9 contains *my*, but the user refers to her child as *clubfoot cutie*. In 10, the user entirely omits an explicit reference to her child, though humans can infer that *termination* is a nominalization of the verb phrase, *terminating our child*.

## 4.3. Limitations

The classification that we presented in this paper is the first step towards scaling the use of social media data for observing pregnancies with birth defect outcomes. Thus, this study is limited in that automatically identifying a cohort for such large-scale epidemiological research depends, furthermore, on automatically (i) determining whether users identified as posting "possible defect" tweets have a child with a birth defect, and (ii) verifying the availability of a user's tweets during the pregnancy with a birth defect outcome. The first task would involve, for example, automatically resolving whether a person's name or personal pronoun (e.g., *she*) in a "possible defect" tweet is referring to the user's child, based on contextual tweets in the user's timelines. For the second task, we will further develop a system [48] for automatically determining the prenatal period and the availability of this timeframe in the timeline of a user with a birth defect outcome.

## 5. Conclusions

In this paper, we presented (i) baseline results of natural language processing (NLP) and supervised machine learning methods for automatically detecting tweets by users reporting their birth defect outcomes, (ii) a comparison of feature-engineered and deep learning-based classifiers trained on imbalanced, under-sampled, and over-sampled data, and (iii) an error analysis that could inform strategies for improving classification performance using the annotated corpus we have made publicly available [23]. In practice, the best-performing classifier will be deployed on unlabeled tweets that are stored in our database [13], automatically pre-filtered through a rule-based module [10], and pre-processed as described in this paper. To advance the large-scale use of social media as a complementary data source for studying birth defects, future work will focus on automating timeline-level analyses—in particular, resolving "possible defect" tweets and detecting the prenatal period.

## Funding

This work was funded by the National Institutes of Health (NIH) National Library of Medicine (NLM) award number R01LM011176. The content is solely the responsibility of the authors, and does not necessarily represent the views of the NIH or NLM.

## Acknowledgements

The authors would like to acknowledge Karen O'Connor and Alexis Upshur for their annotation efforts.